# On Training Sketch Recognizers for New Domains

Kemal Tugrul Yesilbek
Beat Research B.V.
Amsterdam, The Netherlands
`k.yesilbek@thebeat.co`

T. Metin Sezgin
Koc University
Istanbul, Turkey
`mtsezgin@ku.edu.tr`

## Abstract

*Sketch recognition algorithms are engineered and evaluated using publicly available datasets contributed by the sketch recognition community over the years. While existing datasets contain sketches of a limited set of generic objects, each new domain inevitably requires collecting new data for training domain specific recognizers. This gives rise to two fundamental concerns: First, will the data collection protocol yield ecologically valid data? Second, will the amount of collected data suffice to train sufficiently accurate classifiers? In this paper, we draw attention to these two concerns. We show that the ecological validity of the data collection protocol and the ability to accommodate small datasets are significant factors impacting recognizer accuracy in realistic scenarios. More specifically, using sketch-based gaming as a use case, we show that deep learning methods, as well as more traditional methods, suffer significantly from dataset shift. Furthermore, we demonstrate that in realistic scenarios where data is scarce and expensive, standard measures taken for adapting deep learners to small datasets fall short of comparing favorably with alternatives. Although transfer learning, and extensive data augmentation help deep learners, they still perform significantly worse compared to standard setups (e.g., SVMs and GBMs with standard feature representations). We pose learning from small datasets as a key problem for the deep sketch recognition field, one which has been ignored in the bulk of the existing literature.*

## 1. Introduction

Sketching is a natural mode of communication. It is a powerful means of expressing ideas, and conveying concepts [3]. The fast growing body of work on sketch recognition aims to build accurate recognition systems that will eventually enable fluid sketch-based interaction.

Thanks to the advances in feature representation, machine learning, and deep learning in particular, the field has enjoyed substantial growth. There has been significant progress in the development of accurate recognition algorithms. We now have large datasets containing millions of drawings from hundreds of classes. Yet, the final link that brings sketch recognition to real-life applications is still missing due to practical limitations. One such limitation is the requirement to create custom recognizers for each new domain. Each domain comes with a specific set of visual vocabulary, which is unlikely to be covered by the set of categories available from the existing datasets. Hence, one needs to compile a dataset of domain-specific sketches, and train models with this dataset.

This brings about two fundamental concerns: First, will the data collection protocol yield ecologically valid data? Second, will the amount of collected data suffice to train sufficiently accurate classifiers?

The first concern, ecological validity, questions how well the result of an experiment translates to the real life scenario. In the context of sketch recognition, ecological validity requires data collection setup to be as close to the real use case as possible. If, for example, the use case involves sketch-based design, the data should be collected using a protocol that mimics the actual design task. Training on data that is detached from the context and the use case violates ecological validity and is likely to lead to dataset shift followed by inferior recognition performance. The first goal of this paper is to draw attention to this specific issue within a sketch-based gaming use case scenario.

The second concern, sufficiency of data, has the potential to turn into a roadblock in deep-learning-based approaches, which are particularly data-hungry. The use of enormous datasets such as the QuickDraw dataset [14] has become the de facto standard when it comes to evaluating and reporting performance of new methods. As a consequence the characteristics of new methods for small datasets have remained a mystery. Their relative performance compared to other deep learning algorithms, and against more traditional setups not been reported over smaller and more practical datasets. Hence, the second goal of this paper is to advocate reporting performance metrics on small datasets, and more importantly emphasize the need to build deep learning methods that can learn from small datasets.

In order to explore the difficulties posed by these two concerns, we walk through the steps of building a sketch-based application using a gaming use case. More specifically, we scrutinize the decisions that one would have to take over the course of training recognizers for our sketch-based application. First, we describe the process of selecting a set of symbols to be used for sketch-based interaction. Then we describe three alternative data collection protocols for collecting representative data. The protocols have been designed to reflect increasing level of ecological validity ranging from a typical in-lab symbol drawing setup, to an in-situ setup where the data is collected during real game play through a Wizard of Oz setup. Next, we report recognition results obtained under these data collection policies using deep learning methods, and contrast them with a more standard SVM/GBM setup.

We have two main contributions. First, we demonstrate that ecological validity in data collection is a real concern, and one that needs to be directly incorporated into the development of sketch recognition algorithms in order to move beyond upper limits imposed by dataset shift. Second, we show that there is substantial performance discrepancy between what deep sketch recognition methods have demonstrated in large datasets such as QuickDraw, and what they can accomplish with small realistic datasets. This points to a gap in the literature, which we believe, deserves attention of the community.

In the rest of the paper, first we describe the related work under sketch recognition and database collection. Next, we describe the three data collection setups, and discuss the measures we have taken to collect ecologically valid data. In section 4, we evaluate the performance of deep learners and traditional learners trained with the three groups of data with an emphasis on performance loss due to lack of ecological validity. Section 5 explores the data sufficiency concern, and compares the performances of seven deep learners to SVM and GBM learners. Section 6 provides a discussion of the results, and lays out important challenges for the deep sketch recognition community. We conclude with a summary of our contributions and future work.

## 2. Related Work

Although interest in sketch recognition can be traced back to Ivan Sutherland's Sketchpad [28], the field had a revival around the turn of the millennium, thanks to learning-based approaches. Earlier approaches typically combined ink features, image features, temporal features, or their various combinations with conventional classifiers (e.g., SVMs, Bayesian Networks, mixture models) [40] [31] [26] [25] [20]. These approaches were generally developed and tested using relatively small datasets (10-100 examples per class) [5] [32] [1] [18]. Availability of significantly larger set of data enabled superior solutions based on deep learning that provide end-to-end feature extraction and classification.

Today, large sketch datasets consist of millions of examples from hundreds of classes. Two well-known examples are TU-Berlin [9] and QuickDraw [14]. QuickDraw was compiled using a web-based platform that gamifies sketch collection. This dataset has 50,000,000 sketches from 345 classes. The TU-Berlin sketch dataset consists of 20,000 examples across 250 classes. This dataset was collected using Amazon Mechanical Turk, a paid crowd-sourcing service.

The QuickDraw and TU-Berlin datasets enabled researchers to build deep sketch recognition models that achieve impressive results. Using the TU-Berlin dataset, Zhang et. al. [42], and Yu et.al [41] developed CNN-based neural networks for sketch classification. This dataset is also utilized by Sedatti et.al. [23] to develop a the neural network for similarity search. Using QuickDraw dataset, Ha et. al. [10] developed an RNN-based network for sketch generation, Xu et. al. [36] developed a deep hashing method using the mixture of CNN and RNN layers, and Song et. al. [27] developed a network that can map images to sketches using LSTM and CNN layers. Sketch segmentation is another field where deep models contributed. Kaiyrbekov et. al. [13] used RNN and MLP layers, and Yang et. al. [37] used deep graphical neural network to achieve stroke-level sketch segmentation. A recent survey of deep learning methods for sketch data by Xu et. al. [35] offers an extensive overview to the field.

While these massive datasets empowered researchers to achieve state-of-art results using deep learning, the set of object categories they cover is not exhaustive. Hence, in practice, researchers and practitioners are still required to collect data for their custom sketch-based applications or to use existing smaller datasets. For example, Tirkaz et. al. [32] relied on sketch datasets for crisis management and military field operations in their work. Today, the only available datasets for these use-cases are the NICICON [16] and COAD datasets, which are limited in size. A more recent example is the sketch dataset collected by Yesilbek et. al. [39] where the least populated class has only 47 examples. This paper is another example where a sketch dataset had to be collected to develop a custom sketch-based gaming application.

Although there has been efforts to train accurate sketch recognizers using few examples, they are just a handful in number. For example, work by Yesilbek et. al. [39] proposes a self learning framework that can learn from as few as 1-3 labeled examples, however it also requires a large dataset of unlabeled data. We advocate learning from few examples. However our focus is on fully supervised sketch recognition using limited size datasets where unlabeled data is not present.

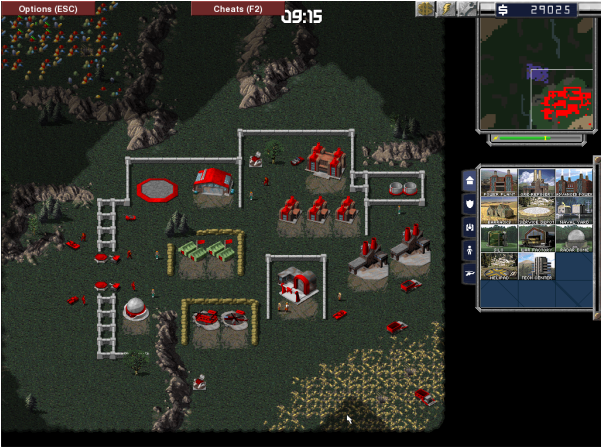

Figure 1. Screenshot from OpenRA, a real-time strategy video game where parties compete in an imaginary military conflict by constructing their bases, producing units, and commanding them. We have modified this open-source game to be played solely with sketching. The base design and implementation offers an optimum environment for intuitive sketch-based interaction.

In recent years, zero-shot sketch-based image retrieval (SBIR) has grasped the attention of sketch recognition community. This field, dominated by deep learning methods, achieved exceptional results by requiring no additional data to perform on novel classes [33] [6] [30] [8] [38]. Unlike SBIR, our focus is on recognition instead of retrieval. The efficacy of zero-shot SBIR methods for recognition tasks has not been assessed.

## 3. Data Collection Protocols

In this section, we first introduce our sketch-based gaming use case. Then we define the three data collection protocols, the adequacy of which we later investigate.

### 3.1. Sketch-based Gaming: A Test-Bed

We picked sketch-based video gaming to serve as a use case. After a survey of possible different genres, we concluded that 2D real-time strategy games would serve as a suitable genre for sketch-based interaction. We picked OpenRA [19], a popular open-source game, and adapted it to support sketch-based interaction. An screenshot of the game is presented in Figure 1.

### 3.2. Designing the set of symbols

The first step in designing a sketch-based interface is to decide on the set of symbols to be used. The field of semiotics is the formal study of signs, symbols and their semantics. We took a semiotics-driven approach to design two sets of symbols for game play.

|  |  | Visual alike | Semi-syntactic |
|---|---|---|---|
| Construction | Power Plant |  |  |
|  | War Factory |  |  |
|  | Turret |  |  |
|  | Naval Yard |  |  |
| Production | Rocket Soldier |  |  |
|  | Ore Truck |  |  |
|  | Light Tank |  |  |
|  | Transport |  |  |
| Selection | Selection |  |  |
| Commanding | Move/Attack |  |  |
|  | Repair |  |  |

Figure 2. Symbol designs considered for our sketch-based video game. Visual alike set has symbols that remind of game object visuals with no syntactic structure. Semi-syntactic set has symbols that are composed of a grammatic composition while still having a visual reminder of the game objects. In this set, building objects have symbols encapsulated with a box, and production object symbols are encapsulated with a circle.

The first set of symbols were designed to resemble loose visual abstractions of the objects used in the game. We refer to this set as the visually alike set. The second set we designed aimed to achieve weak grammatic compositionality in symbol design, where symbols have parts to refer their abilities. This is the approach that has been taken to design the course of action symbology that is actually used in the US armed forces, formally defined in the operational terms and graphics manual [5]. We refer to this second set of symbols as the semi-syntactic set. We present these sets in Figure 2.

### 3.3. Sketch Data Collection Setups

Research in fields such as face and object recognition have transitioned to realistic datasets [12][34]. However, the sketch recognition community, and particularly the deep sketch recognition community, has primarily been concerned about the scale and open access availability of datasets. Data quality and validity have largely been neglected [17]. It is easy to encounter examples of sketch recognizers trained on datasets collected in setups that mismatch to their use case. NICICON [16] is an example. Although the NICICON dataset was collected to build recognizers to serve in emergency management, it was collected in an isolated lab-like setup. The imperative nature of emergency situations are likely to distort sketches which may result in poor recognition performance. Collecting sketch data in a lab-like setup ignores this factor, makes it impos-

| Testing Set: | Realistic | | | Pseudo-Realistic | | | In-Lab | | |
|---|---|---|---|---|---|---|---|---|---|
| Training Set: | R | P-R | I-L | R | P-R | I-L | R | P-R | I-L |
| SVM | 98.7% | 94.0% | 94.6% | 89.4% | 96.1% | 93.6% | 95.0% | 97.1% | 97.2% |
| GBM | 94.6% | 89.2% | 89.6% | 90.0% | 93.6% | 91.5% | 93.5% | 96.5% | 96.6% |
| CNN-L | 88.1% | 79.9% | 83.3% | 84.6% | 86.9% | 86.1% | 87.6% | 88.5% | 91.7% |
| CNN-M | 88.4% | 79.8% | 83.4% | 88.4% | 86.3% | 86.1% | 88.4% | 90.0% | 91.5% |
| CNN-S | 91.5% | 83.3% | 85.7% | 87.1% | 87.9% | 88.3% | 90.5% | 92.5% | 93.2% |
| CNN-XS | 91.7% | 83.0% | 86.4% | 87.1% | 88.4% | 87.6% | 91.6% | 93.2% | 93.7% |
| EM-L | 88.7% | 82.1% | 80.0% | 77.0% | 85.6% | 79.8% | 69.0% | 73.3% | 79.9% |
| EM-M | 90.1% | 82.9% | 82.1% | 77.8% | 86.2% | 81.5% | 68.5% | 73.7% | 80.3% |
| EM-S | 89.3% | 83.6% | 79.4% | 75.9% | 85.0% | 79.6% | 68.6% | 72.8% | 78.1% |

Table 1. Performance experiment results with different train-test combinations. Experiments are conducted with standard 10 fold cross-validation. Results are reported in mean accuracy. Yellow shaded cells highlights the best method, and blue shaded cells highlights the best deep learning method for the column. The results clearly shows that models trained using protocols that lacks in ecological validity performs poorly in real-life use. Another insight is that deep learning models cannot reach to the performance level of more traditional learning methods in small datasets. R: Realistic, I-L: In-Lab, P-R: Pseudo-Realistic

sible to evaluate real-life performance of the recognizer.

We collected three different datasets to study the effects of the mismatch between the collection and real use environment for sketch data. Datasets are collected via protocols designed to reflect increasing level of ecological validity.

Our first dataset serves as a baseline. This dataset was collected in an isolated, lab-like setup. Participants of this data collection activity were simply asked to sketch displayed symbols on a tablet using a stylus. This dataset represents the status-quo way of collecting sketch data. We will refer to this dataset as **in-lab** in the rest of this paper.

The second dataset is an intermediary between an in-lab and a realistic setup. While this dataset was also collected in an isolated environment, participants are put under time-pressure to mimic the sense of urgency players have when competing in a real-time strategy game. To create the time-pressure, we modified our data collection software to award scores to each drawn symbol inversely proportional to its sketching duration. We instructed the participants to beat previous participants by aiming to achieve maximum score. Our main goal with this intermediary setup was to assess the plausibility of mimicking a realistic setup while keeping the low-cost structure of an in-lab setup. We refer to this dataset as **pseudo-realistic** in the rest of this paper.

The third dataset is collected in an in-situ setup that has complete ecological validity. This fully realistic dataset was collected while participants played our sketch-based game. Collecting this dataset had two major challenges. The first challenge was to have the sketch-based game fully functional. This required us to replace the standard mouse-keyboard setting with sketch-based interaction. The second challenge was the absence of a custom sketch recognizer. Since a custom high performant recognizer was not present, we had to establish a mechanism to substitute it. We overcame this challenge by assembling a Wizard of Oz [4] setup. In this setup, the sketches drawn by participants are recognized by an Oracle. Sketches and labels are exchanged between the participant and the Oracle's computer in real-time, providing the illusion of a perfect recognizer in operation. Participants of this data collection activity were asked to compete against an AI-controlled player after a short tutorial. Choosing a popular game was helpful in this setting as all of the participants were familiar with the game mechanics. Participants freely competed against the AI player while we collected sketches. The participants were not informed about the Wizard until after the end of the study.

## 4. Performance Evaluations Comparing Data Collection Setups

We ran experiments to measure how models trained with datasets collected under different protocols perform in a real-life scenario. We considered both statistical and deep learning methods for our experiments. We selected GBM and RBF-Kernel SVM as a representative subset of statistical learning methods. For deep learning, we considered state-of-art CNN-based and embedding-based architectures of various sizes.

We used IDM [21] as the feature extraction method for statistical learning. For many years, IDM served as the state of art feature extraction method for sketch data before the wide adoption of deep learning by the sketch recognition community. The hyper-parameter search for SVM was car-ried out using the intervals suggested by Yesilbek et. al. [40]. Our GBM models were trained with the CatBoost implementation [7]. We kept the default hyper-parameter values set by the library across our experiments. This flavour of GBM has reported the highest performances with default hyper-parameters compared to its alternatives.

There are numerous approaches to deep learning for sketch recognition. In this paper, we focused on CNN and embedding based methods. We evaluated four architectures based on CNNs, and three architectures based on embedding extraction. For each method, we employed architectures varying in the number of parameters. CNN-based architectures were derived from the architecture proposed in sketch-a-net paper by Yu et. al. [41]. We label these architectures as CNN-L, CNN-M, CNN-S, and CNN-XS referring to their size. The CNN-L architecture is implemented by following the sketch-a-net paper, which has 8,502,473 parameters. Following a similar setup, remaining architectures have reduced number of layers and filters. The CNN-M architecture has 537,481, CNN-S architecture has 79,401, and CNN-XS architecture has 23,577 parameters. Embedding-based architectures evaluated in this paper implement the embedding extraction segment proposed in Sketch Former by Riberio et. al. [22]. The embedding extraction segment is trained with the full size QuickDraw dataset. This segment maps a given sketch into a 128 length vector ready to be used for various purposes. In order to achieve classification from the embedding vectors, we adopted three different fully connected feed-forward neural networks with varying in size. We label these architectures as EM-L, EM-M, and EM-S referring to their sizes. Excluding the embedding segments, the EM-L architecture has 260 units, followed by EM-M with 140 units, and EM-S with 112 units.

We conducted 81 experiments in total using nine methods to get a complete picture. In each experiment, a model is trained and tested with different datasets collected using different protocols in varying ecological validity. We present our results in Table 1. Statistical significance analysis using multi-factor ANOVA shows significant difference interaction for $p < 0.05$ for factors of architecture, training set, and test set. The data used in these analyses have satisfy ANOVA's assumptions (normality, homoscedasticity, and no multicollinearity).

Our first conclusion from the results is that the level of ecological validity has a significant impact on the recognizer performance. In all choices of classifier, we observed that the models trained with the in-lab data have failed to learn the sketching patterns that emerged in the realistic setup. This indicates that a low level of ecological validity causes substantial dataset shift which deteriorates the recognizer performance in a real life scenario. As an example, the practitioner who neglect ecological validity in their data collection setup would report 91.7% accuracy for their model with CNN-L architecture. However this model would only have performed with 83.3% accuracy in the real-life use-case. In the light of this conclusion, we invite sketch recognition practitioners and researchers to collect their datasets in an in-situ setup to avoid the dataset shift. We advocate reporting performance results using a dataset collected with an in-situ setup since it gives a more accurate estimation of how the models will perform in real life scenarios.

We also conclude from the results that using protocols attempting to mimic the in-situ setup is not sufficient. Experiments show that training models with pseudo-realistic data fails to improve the recognizer performance when tested with realistic data compared to models trained with in-lab dataset. Furthermore, the small gap between the models trained with in-lab and pseudo-realistic data indicates that our pseudo-realistic setup failed to generate sketches with patterns similar to a realistic setup. Based on this result, we conclude that mimicking the in-situ setup requires rigorous study of the effects and constraint by the real life scenarios.

## 5. The Effect of Small Data over Performance

As illustrated in our sketch-based gaming use case, moving to a new domain inevitably requires collecting realistic examples of domain specific symbols. In our case, this meant an exhausting effort to build the infrastructure collecting in-situ data, recruiting and compensating participants willing to play the game, running the experiment in presence of an experimenter and a wizard, and a follow up effort for data labeling. This amounted to 2 person-months of labor for building the data collection infrastructure, and 2 days for data collection with 10 participants. Even then, we were only able to collect a modest amount of data as seen in Table 2.

Our dataset is orders of magnitude smaller than the large datasets which have become standard in the field for developing and evaluating recognition algorithms. In contrast to the vast number of work demonstrating the superiority of deep learners, Table 1 shows that statistical learning methods have a clear edge on small datasets. This, in itself, is not surprising as deep learners claim superiority on big data, but make the remark that strategies involving data augmentation and the use of pretrained models can be deployed for small datasets.

In accordance, we assess the efficacy of these methods for our use case using extensive data augmentation and transfer learning. We compare the results to those obtained using SVMs and GBMs, again using the standard IDM feature represenation [21].

Transfer learning is widely adopted in deep learning field [29] [2] [15]. This approach promotes initializing model parameters from pre-trained models instead of setting them randomly. It has been shown that sketch recognizers bene-fit from this technique similar to other fields [24] [11]. We adopted this technique aiming to achieve better results with our deep learning model. We initialized the parameters of the model with values learned by pre-training. The pre-training was based on 14 randomly selected classes from the QuickDraw dataset. This allowed the model to have a head start in training. As our results show, initializing model parameters with a pre-trained model increases the performance of our model.

| Type | Class | Number of Examples |
|---|---|---|
| Object | Gunboat | 52 |
| Object | Turret | 61 |
| Object | Rocket Soldier | 61 |
| Object | Pillbox | 74 |
| Object | Power Plant | 83 |
| Object | Rifle Infantry | 87 |
| Object | Light Tank | 106 |
| Action | Do | 122 |
| Action | Select | 122 |

Table 2. Number of examples collected for each class in our realistic dataset. Low example count and non-uniform distribution is an artifact of unrestricted data collection setup.

| | Training Dataset | | |
|---|---|---|---|
| Method | Realistic | Pseudo-realistic | In-lab |
| SVM | 98.7% | 94.0% | 94.6% |
| GBM | 94.6% | 89.2% | 89.6% |
| EM-M | 90.1% | 82.9% | 82.1% |
| CNN-S | 92.3% | 83.5% | 85.4% |
| CNN-S + Transfer L. | 94.0% | 87.6% | 89.4% |
| CNN-S + Transfer L.+ Data Aug. | 95.1% | 90.1% | 90.6% |

Table 3. Mean accuracy results with additional techniques applied to CNN-S model. All models are tested with the Realistic dataset. Data augmentation generated 50 new examples for each sample. Results show that even though deep learning models improve with additional techniques trained with small dataset, their performance still can't reach to traditional learning methods.

As second remedy, we performed data augmentation. Applying rotation and shearing transformations to existing examples, we synthetically increased our dataset size by a factor of 50. When applied in combination with transfer learning, this technique further improved the performance of our deep learning model.

Table 3 shows the results obtained using transfer learning, extensive data augmentation and their combination. The results demonstrate that transfer learning in combination with data augmentation significantly improved the performance of our deep learning model without requiring us to obtain additional examples. While still falling short compared to more traditional methods, we show that deep learning methods can achieve reasonable results on small datasets by employing additional techniques.

## 6. Conclusion and Summary

The effect of ecological validity in sketch datasets was previously unexplored. In order to investigate how data collection setups effect sketch recognition, we collected datasets in setups with varying levels of ecological validity. Our sketch-based gaming use case paired with Wizard of Oz setting allowed us to collect a fully realistic dataset.

We conducted experiments using various learning methods including more traditional ones like SVM, and GBM, as well as deep learning models varying in approach and size. Our results show that data collection protocols have significant impact on model performances. Moreover, we showed that training models with a dataset collected in an isolated environment causes dataset shift, which results in lower recognition performance in real-life use. Our attempts to mimic the in-situ setup has failed to show better results when compared to in-lab setting indicating that users' sketching behavior change in unobvious when using the application.

Collecting sketch data in a realistic setup can be very expensive. This high-cost structure prevents assembling large realistic datasets. This is a serious limitation for deep methods. In our experiments, deep learning methods performed significantly worse than the more traditional ones regardless of their size or approach. Even with transfer learning and substantial data augmentation, their accuracy was subpar.

In conclusion, we invite sketch recognition practitioners to collect their data using an in-situ setups to avoid the dataset shift. Further, we urge researchers to report results with smaller datasets along with larger ones as they try to push beyond the state of the art.

## 7. Future Work

Transfer learning, and data augmentation taken individually and together failed to achieve substantial improvements in recognition accuracy. We believe this area deserves the attention of the community. Style transfer in particular could be a fruitful direction to explore. For example, transferring drawing style from few in-situ examples to large in-lab data could serve as a novel means of generating realistic in-situ-like data.

## 8. Acknowledgements

The authors gratefully acknowledge the support of KUIS AI Center, TUBA, and Bilim Akademisi.